\documentclass{article}
\usepackage{spconf,amsmath,graphicx}
\usepackage{url}
\usepackage{booktabs}
\usepackage{threeparttable}
\usepackage{multirow}

\title{Probabilistic Inference for Camera Calibration in Light Microscopy under Circular Motion}
\name{Yuanhao Guo$^1$, Fons J. Verbeek$^3$, Ge Yang$^{1,2}$}
\address{$^1$ Computational Biology and Machine Intelligence Group, National Laboratory of Pattern Recognition, \\
 Institute of Automation, Chinese Academy of Sciences, Beijing, China\\
$^2$ School of Artificial Intelligence, University of Chinese Academy of Sciences, Beijing, China\\
$^3$ Bioinformatics and Imaging, LIACS, Leiden University, Leiden, Netherlands}
\begin{document}

%
\maketitle
\begin{abstract}
Robust and accurate camera calibration is essential for 3D reconstruction in light microscopy under circular motion. Conventional methods require either accurate key point matching or precise segmentation of the axial-view images. Both remain challenging because specimens often exhibit transparency/translucency in a light microscope. To address those issues, we propose a probabilistic inference based method for the camera calibration that does not require sophisticated image pre-processing. Based on 3D projective geometry, our method assigns a probability on each of a range of voxels that cover the whole object. The probability indicates the likelihood of a voxel belonging to the object to be reconstructed. Our method maximizes a joint probability that distinguishes the object from the background. Experimental results show that the proposed method can accurately recover camera configurations in both light microscopy and natural scene imaging. Furthermore, the method can be used to produce high-fidelity 3D reconstructions and accurate 3D measurements.
\end{abstract}
\begin{keywords}
Camera calibration, circular motion, light microscopy, probabilistic model, 3D reconstruction
\end{keywords}

\begin{figure*}[!t]
\centering
\includegraphics[width=0.85\textwidth]{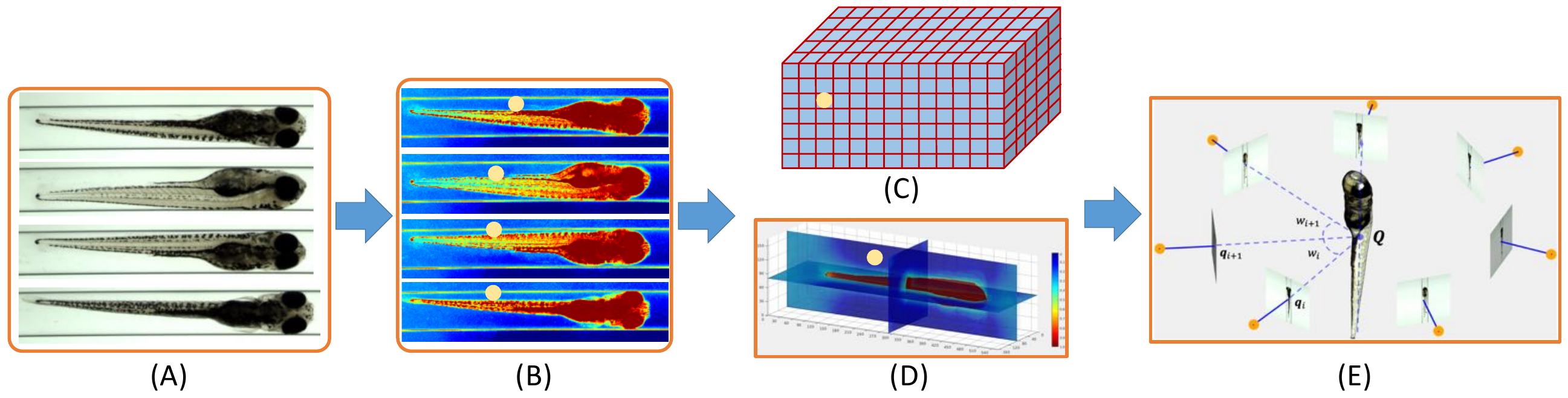}
\caption{A graphical scheme of the proposed method. (A) The images acquired from light microscopy under circular motion. (B) The probabilistic representation of the images. The warmer the color of a pixel is, the more likely it is classified as foreground. (C) A set of shape constrained voxels. The yellow dots track the matching from the voxel to the images. (D) The confidence map obtained from the integration of all the images. (E) The probabilistic inference for the camera calibration.}
\label{fig:fig1}
\vspace{-0.2cm}
\end{figure*}

\vspace{-0.2cm}
\section{Introduction}
\label{sec:intro}
\vspace{-0.2cm}

Circular motion is commonly used in light microscopy because it simplifies the imaging protocol \cite{pardo2010high}. A specimen in circular motion presents its longitudinal views from a full revolution, which makes most of its features observable. The acquired images can be considered as the readout of multiview imaging. Thus, the image-based 3D reconstruction approaches may produce a 3D volumetric representation of a specimen, from which 3D visualization and quantitative measurements can be performed. Because this type of methods requires an accurate estimation of the projection from the 3D world frame to each of the images, a robust and accurate camera calibration becomes necessary \cite{hartley2003multiple,ben2016camera}, even for the popular deep learning based reconstruction methods \cite{kar2017learning,huang2018deepmvs,zhang2018learning}.  

Conventional calibration methods capture images for a regular pattern like a checkboard to estimate the camera intrinsic configurations, including the focal length, pixel size, image center, etc \cite{heikkila1997four,zhang2000flexible}. Bundle adjustment is then taken to get the camera extrinsic configurations, i.e. the camera motion (transformation) against the subject \cite{wu2011multicore}. Alternatively, the silhouette-based methods achieves this goal by taking the multiview masks of the subject as input. Three-dimensional (3D) projective geometry generates an accurate projection from the subject to each of the masks. The area coherence based method aims to maximize the overlap area between the projected images and the groundtruth masks \cite{lensch2001silhouette}; The silhouette coherence based method aims at maximizing the intersection of the contours from the projected images and the groundtruth masks \cite{hernandez2007silhouette}; The shape coherence based method defines the voxel residual as the intersection of the cone-shaped projections, thus avoiding extract contours or surface area of the projected images \cite{guo2017three,guo2017two}. It should be noted that the methods mentioned above depend on either accurate key point matching or precise image segmentation. However, both are often difficult to achieve in light microscopy due to the commonly existing transparency/translucency of specimen. 

To address the above issues, we propose a probabilistic inference based method for the camera calibration. In contrast to previous methods, our model is advantageous because it does not require accurate key point detection or precise image segmentation. Our method first transforms the images acquired from the circular motion into a probabilistic representation without the need for image segmentation. We then initialize a range of voxels which may cover the subject of interest (SoI). We project those voxels onto the probabilistically represented images according to 3D projective geometry. In such a manner, we assign a joint probability on each voxel by integrating the information from all the axial-view images. Finally, we formulate the camera calibration problem as the maximization of the separation of the SoI from the background, which is represented as a probabilistic inference. Through this probabilistic inference, we do not need to assign a hard-coded value on each voxel, which is usually determined by its total visibility from all the segmentations (silhouettes). We demonstrate the proposed method in a graphical scheme shown in Fig. \ref{fig:fig1}.

We organize the remainder of this work  as follows. In Section \ref{sec:method} we elaborate the proposed method. In Section \ref{sec:experiments} we show the experimental results and the performance of our method. In Section \ref{sec:conclusions} we summarize our work and suggest some directions for future improvement of our method.

\vspace{-0.3cm}
\section{Methodology}
\label{sec:method}
\vspace{-0.1cm}
In this section, we will elaborate the proposed probabilistic inference for the camera calibration in light microscopy under circular motion. The zebrafish are commonly used in life-science study \cite{howe2013zebrafish}. The images used in this study were collected from a group of zebrafish larvae using a light microscope under circular motion based on the VAST-BioImager \cite{pardo2010high}. The VAST-BioImager loads a specimen into the view of a high-throughput camera (Allied Vision Systems, Pro Silica GE 1050 CCD, pixel size $5.5 \mu m \times 5.5 \mu m$, image size $1024\times 1024$ pixels). A stepper motor manipulates the specimen to rotate along its longitudinal axis and the equipped camera captures image for each axial-view. Some examples of the acquired images are shown in Fig. \ref{fig:fig1} (A).

\vspace{-0.3cm}
\subsection{Camera Parameterization}
\vspace{-0.1cm}
\label{subsec:camera_param}
Let $\mathbf{Q}=(x,y,z) \in R^3$ denote the center of a voxel in 3D world frame. The classical pinhole model finds its pixel location $\mathbf{q}$ on an image via the projection matrix $\mathbf{P} \in R^{3 \times 4}$, denoted as $\mathbf{q}=\mathbf{P} \cdot \mathbf{Q}$. $\mathbf{P}$ can be decomposed as $\mathbf{P}=\mathbf{K}[\mathbf{R} | \mathbf{t}]$. $\mathbf{K}$ is the camera intrinsic matrix of the following form.

\vspace{-0.2cm}
\begin{equation}
\label{eq:matrix_K}
\mathbf{K} = \left[
\begin{array}{ccc}
f\cdot k_x & 0 &  \mu_x \\
0 & f\cdot k_y &  \mu_y \\
0 & 0 & 1 \\
\end{array}
\right]
\end{equation}

Where, $f$ denotes the focal length; $(k_x, k_y)$ denote the scaling factors; $(\mu_x,\mu_y)$ define the image center. Because these configurations are provided by the vender, we focus on the camera extrinsic parameters, i.e. the camera motion, encoded in $\mathbf{R}$ and $\mathbf{t}$. Those two terms are the 3D rotation and translations, respectively. As the subject in our settings is rotating along a fixed axis, the yaw and pitch remain identical during motion; the roll and translation vary from different views. So, we organize the camera extrinsic parameters that will be optimized into a vector $\theta=[\alpha, \phi, \gamma, \omega_{1:N-1}]$, where $\alpha$ and $\phi$ are the yaw and pitch; $\gamma$ is the reparameterized translation; $\omega_{1:N-1}$ include the roll for each view (the starting view is defined as 0 degree). Given a fixed camera configuration $\theta$,  we can derive the projection matrix $\mathcal{P}=\{\mathbf{P}_1, \mathbf{P}_2, \cdots, \mathbf{P}_N \}$ for each image.

\vspace{-0.2cm}
\subsection{Probabilistic Representation}
\label{subsec:prob_repre}
\vspace{-0.1cm}
According to the color distribution, we construct two probabilistic models for the foreground ($f$) and background ($b$). We randomly select one image from the axial-view images. Then we either scratch on the image or roughly mask the image to collect the pixels separating the foreground and background. According to the normal distribution model, we use the RGB values of the collected pixels to construct the probabilistic models as follows.

\vspace{-0.5cm}
\begin{equation}
\label{eq:prob_model_shape}
\begin{aligned}
&p_f(\mathbf{q}) = \frac{1}{\sqrt{(2\pi)^3|\mathbf{\Sigma}_f|}}e^{-\frac{1}{2}(\mathbf{I}(\mathbf{q})-\mathbf{\mu}_f)^T\mathbf{\Sigma_f ^{-1}} (\mathbf{I}(\mathbf{q})-\mathbf{\mu}_f)}\\
&p_b(\mathbf{q}) = \frac{1}{\sqrt{(2\pi)^3|\mathbf{\Sigma}_b|}}e^{-\frac{1}{2}(\mathbf{I}(\mathbf{q})-\mathbf{\mu}_b)^T\mathbf{\Sigma_b ^{-1}} (\mathbf{I}(\mathbf{q})-\mathbf{\mu}_b)} ,
\end{aligned}
\end{equation} 
\vspace{-0.2cm}

Where $\mu$ and $\mathbf{\Sigma}$ denote the mean and covariance of the color distribution, respectively. Given the $N$ axial-view images $\mathcal{I}=\{\mathbf{I}_1, \mathbf{I} _2, \cdots, \mathbf{I}_N\}$, we obtain the probabilistic representation for each of the images based on Eq.\ref{eq:prob_model_shape}.

\vspace{-0.2cm}
\subsection{Probabilistic Inference for Camera Calibration}
\label{subsec:camera_calib}
\vspace{-0.1cm}
Given a voxel $Q_j$, $j\in[1,M]$, and a projection matrix for an image $\mathbf{P}_i$, $i\in [1,N]$, we could find the pixel location of $Q_j$ on that image through $\mathbf{q}_{i,j}=\mathbf{P}_i\cdot\mathbf{Q}_j$. In this way we can match the pixel location for the voxel $Q_j$ on each of the image by the $\mathbf{q}_{\cdot,j}$.  By integrating the probabilities of the voxel $Q_j$ from all the $N$ images, we obtain its total probabilities which denote its likelihood of belonging to the foreground and background. This may be implemented by the form of joint probability distribution \cite{kolev2012fast}.

\vspace{-0.2cm}
\begin{equation}
\label{eq:multi_prob}
\begin{aligned}
& P_f(\mathbf{Q}_j) = \left( \prod_{i=1:N} p_f(\mathbf{q}_{i,j})\right)^{\frac{1}{N}}\\
& P_b(\mathbf{Q}_j) = 1 - \left(\prod_{i=1:N}(1 - p_b(\mathbf{q}_{i,j})\right)^{\frac{1}{N}}
\end{aligned}
\end{equation}
\vspace{-0.2cm}

Taking all the $M$ voxels $\mathcal{Q}=\{\mathbf{Q}_1, \mathbf{Q}_2, \cdots, \mathbf{Q}_M\}$ into account, we formulate the cost function as the joint probability distribution of all the voxels, which indicates the total likelihood of the separation from the foreground to the background in the constrained 3D world frame.

\vspace{-0.2cm}
\begin{equation}
\label{eq:cost_func}
\mathcal{L(\theta)}=\prod_{j=1:M}\frac{P_f(\mathbf{Q}_j)}{P_b(\mathbf{Q}_j)}
\end{equation}
\vspace{-0.2cm}

For simplification, we use the following logarithmic form of the cost function.

\vspace{-0.4cm}
\begin{equation}
\label{eq:max}
\begin{aligned}
\theta^*&=\mathop{\arg\max}_{\theta\in\mathcal{\theta}}\mathcal{L(\theta)}\\
&=\mathop{\arg\max}_{\theta\in\mathcal{\theta}}\sum_{j=1:M}[\log(P_f(\mathbf{Q}_j))-\log(P_b(\mathbf{Q}_j))]
\end{aligned}
\end{equation}
\vspace{-0.2cm}

\begin{figure}[!t]
\centering
\includegraphics[width=0.85\columnwidth]{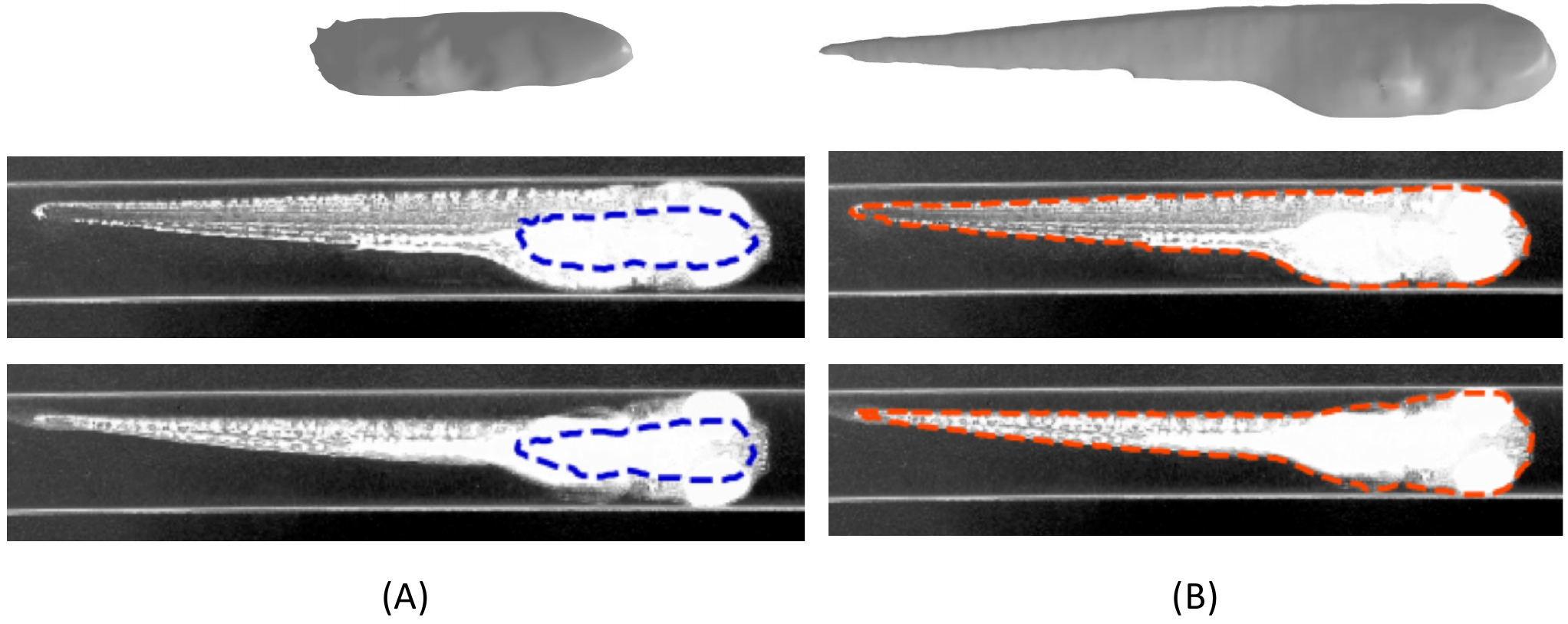}
\caption{An illustration of the effect of camera calibration.  (A) A noisy camera configuration generates a poor 3D model and the projections departure from the realm of the subject.  (B) A well-calibrated camera configuration produces a vivid 3D model of which the projected masks tightly attach the subject. }
\label{fig:fig2}
\vspace{-0.2cm}
\end{figure}

We use the evolution strategy\cite{hansen2004evaluating}, one of the unconstrained optimization methods, to maximize the cost function. We have found that the evolution strategy offers a stable and robust performance if given a reasonable initialization. Specifically, we initialize the camera parameters as $\theta_0=[0,0,0,\Delta\omega\cdot(1:N-1)]$, where $\Delta\omega$ is the initial circular motion step manipulated by the stepper motor. We terminate the optimization when the output of the cost function is smaller than $10^{-6}$. In Fig.\ref{fig:fig2}, we compare the reconstruction effect of the camera calibration. A well-calibrated camera configuration produces a high-fidelity 3D model, and the projected masks can well-distinguish the subject. 

\begin{figure*}[!t]
\centering
\includegraphics[width=0.75\textwidth]{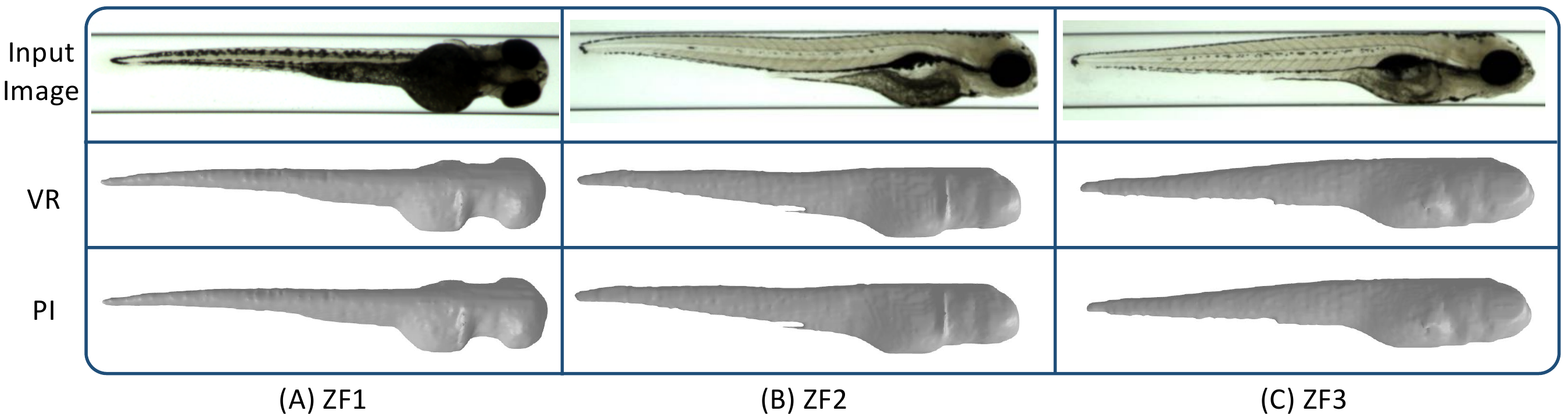}
\caption{Reconstructed 3D models for the zebrafish dataset with different calibration methods.}
\label{fig:fig3}
\end{figure*}

\begin{figure*}[!h]
\centering
\includegraphics[width=0.75\textwidth]{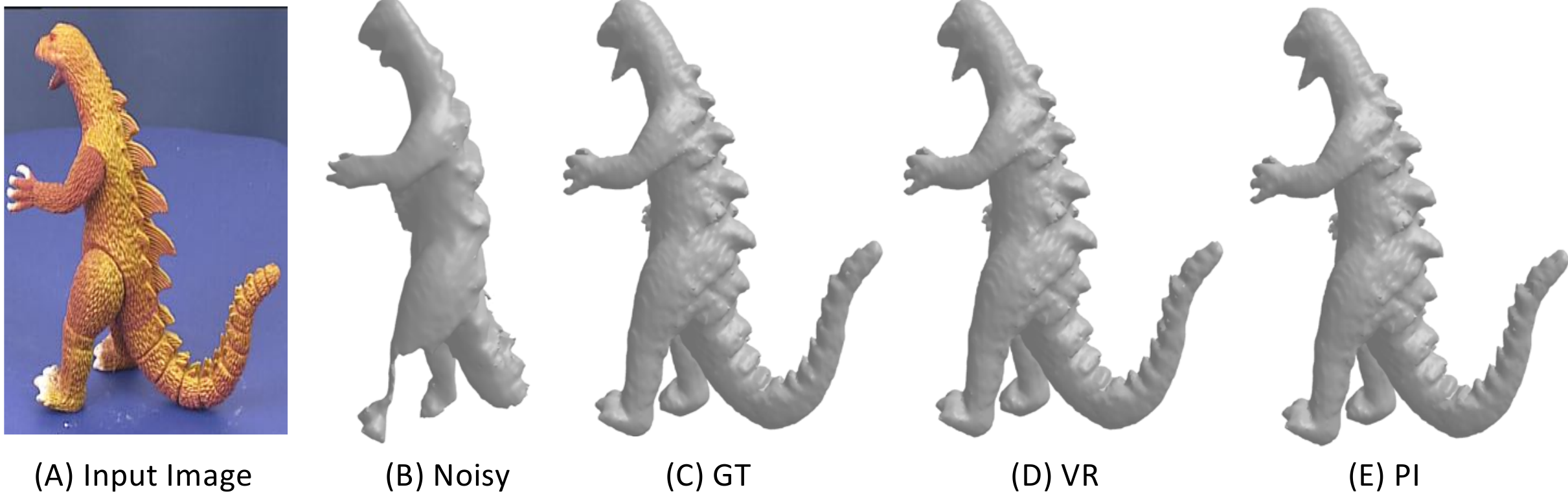}
\caption{Reconstructed 3D shapes for the dinosaur statue with different calibration methods.}
\label{fig:fig4}
\vspace{-0.5cm}
\end{figure*}

\vspace{-0.2cm}
\section{Experiments}
\label{sec:experiments}
\vspace{-0.2cm}
In this section, we set up the \textbf{v}oxel \textbf{r}esidual (VR) maximization method \cite{guo2017three} as a baseline. We separately evaluate the VR and the proposed \textbf{p}robabilistic \textbf{i}nference (PI) model on a dataset acquired from our light microscope imaging system with circular motion. In order to evaluate the generalizability of our method, we compare the performance of the two methods on a public dataset of a dinosaur statue which is vailable from \url{https://www.robots.ox.ac.uk/~vgg/data/data-mview.html}. The pythonic implementation of our method is available at \url{https://github.com/yuanhaoguo/circular-camera-calibration}. We transfer some functions from \cite{BTordoff2016}.
\vspace{-0.2cm}
\subsection{Performance on Light Microscopy Data}
\label{subsec:experiment1}
\vspace{-0.1cm}
We acquired images for three zebrafish specimens (ZF1, ZF2, ZF3) separately from 3, 4 and 5 days post fertilization (DPF). For each specimen, we evenly sample 21 views from the full revolution. The mechanical drift makes the rotation step of the specimen unstable and the rotation axis is not aligned with the specimen center. Thus, we need to calibrate the camera configurations to obtain accurate 3D models of the specimens.

As we do not have the \textbf{g}roud\textbf{t}ruth (GT) camera parameters in this dataset, we propose two strategies to compare the performance of the VR and PI methods. Based on the calibrated camera data obtained from these two methods, we first apply the shape-based 3D reconstruction method\cite{guo2017three} to obtain 3D models. We then (1) inspect the visual effects of the 3D shapes; (2) use two 3D metrics, i.e. \textbf{v}olume (V) and \textbf{s}urface \textbf{a}rea (SA), to measure the 3D shapes.

In Figure \ref{fig:fig3} we visualize one selected view of the 3D reconstructed zebfafish models. We find that the proposed PI model provides similar performance as the baseline method, but without the need for accurate image segmentation. In Table \ref{tab:tab1} we show the 3D metrics. The VR method has been validated for its performance in obtaining accurate 3D measurements in light microscopy imaging under circular motion \cite{guo2017three}. We see that the 3D metrics of the PI model are very close to those of the VR method. In fact, our method applies in diverse types of light microscopes, e.g. the bright-field and fluorescent, and diverse types of biological models, e.g. the zebrafish and mice (data not shown). 

\vspace{-0.2cm}
\begin{table}[!h]
\renewcommand{\arraystretch}{0.85}
\centering
\caption{3D Metrics Comparison on Zebrafish Data}
 \begin{threeparttable}[h]
\begin{tabular}{c|cccc}
\toprule
 & & ZF1 & ZF2 & ZF3 \\
 \midrule
 \multirow{2}{*}{V($mm^3$)} & VR & 0.251 & 0.281 & 0.326 \\
                                          & PI & 0.249  & 0.278 & 0.325 \\ \hline
                                          
 \multirow{2}{*}{SA($mm^2$)} & VR &3.25 & 3.57 & 3.88 \\
                                                   & PI & 3.24 & 3.57 &  3.87 \\                                    
\bottomrule
\end{tabular}
\end{threeparttable}
\label{tab:tab1}
\vspace{-0.5cm}
\end{table}

\vspace{-0.2cm}
\subsection{Performance on Public Dataset}
\label{subsce:experiment2}
\vspace{-0.1cm}
For the public dataset, i.e. the dinosaur statue, we have the \textbf{g}round\textbf{t}ruth (GT) calibration data. Since its camera parameterization is slightly different from the settings in our zebrafish dataset, we design a strategy to enable the evaluation of the VR and PI methods on this dataset. (1) We decompose the camera parameters from the calibration data. (2) We keep the camera intrinsic parameters fixed, and add a certain level of noise on the camera extrinsic parameters, aligning the evaluation to our circular motion imaging protocol. Specifically, we choose 10 rotation angles of roll from the total 36 ones. We perturb the settings by adding 10 degrees on each of the chose angles. (3) We apply the VR and the proposed PI methods to optimize those noisy camera parameters. (4) We compare the optimized camera parameters with the GT data.  

In Table \ref{tab:table2} we compare the calibrated camera parameters from the VR and PI methods with the GT. Our method obtains rather accurate camera calibration result in the natural scene imaging condition. For a perceptual evaluation, we again apply the shape-based method to reconstruct the 3D model of the dinosaur statue. In Figure \ref{subsec:experiment1} (A), we show one original input image; We then separately show the 3D model reconstructed from: (B) the noisy camera parameters; (C) the groundtruth camera parameters; (D) the calibrated camera parameters of the VR method; (E) the calibrated camera parameters of the proposed PI model. The results show that our method can be used for the generic multiview stereo.

\begin{table}[!t]
\renewcommand{\arraystretch}{0.8}
\caption{Calibration Results Comparison on Public Data}
\centering
 \begin{threeparttable}[h]
\begin{tabular}{ccccccccccc}
\toprule
Roll & $\omega_1$ &  $\omega_2$ &  $\omega_3$ &  $\omega_4$ &  $\omega_5$  \\
\midrule 
GT & 87.25 &  79.00 & 69.15 &  59.22 & 49.21 \\
VR & 87.25 &  78.95 & 69.01 &  59.11  & 48.77 \\
PI & 87.29 &  78.88 & 69.18 & 59.22  & 48.91 \\
\toprule
Roll &  $\omega_6$ &  $\omega_7$ &  $\omega_8$ &  $\omega_9$ &  $\omega_{10}$ \\
\midrule
GT & 39.21  & 29.23 & 19.28 & 9.28 & -1.36 \\
VR & 38.40 & 28.28 & 18.25 & 8.55 & -1.26  \\
PI & 38.50  & 28.66 & 18.59 & 8.76 & -1.24 \\
\bottomrule&
\end{tabular}
\end{threeparttable}
\label{tab:table2}
\vspace{-0.6cm}
\end{table}

\vspace{-0.2cm}
\section{Conclusions}
\label{sec:conclusions}
\vspace{-0.2cm}
We proposed a new camera calibration method for light microscopy under circular motion. Our method takes a probabilistic inference to maximize the likelihood for the separation of the foreground and background. The method is free to complicated image pre-processing, like key point detection and image segmentation, which still recovers accurate camera parameters in light microscopy imaging.The potential of the method lies in its successful application in the camera calibration of generic imaging situations. Here we point out two issues. (1) As the cost function optimization in our method is unconstrained, which requires a relatively good initialization for the solution. In most imaging setups the mechanical parameters are usually known, e.g. the rotation angle of the control motors, so this problem could be solved by offering a reasonable estimation for the camera extrinsic parameters. (2) Due to the intensive 3D projections from the voxels to each of the images, the efficiency of our method still needs to improve. We plan to apply a parallel scheduler to accelerate the performance of the proposed method.

\vspace{-0.5cm}
\section*{\large Acknowledgement}
\vspace{-0.3cm}
This study is supported in part by grant from the Chinese Academy of Sciences (No.Y9S9MS01/292019000056) and grant from the University of Chinese Academy of Sciences (No.115200M001). This study is partially supported by the National Natural Science Foundation of China (No.31971289). We thank Dr. Wolfgang Niem at Univ of Hannover and VGG group at Univ of Oxford to provide the dinosaur data. 
\bibliographystyle{IEEEbib}
\bibliography{strings,refs}

\end{document}